\documentclass[a4paper]{article}

\usepackage[pages=all, color=black, position={current page.south}, placement=bottom, scale=1, opacity=1, vshift=5mm]{background}

\usepackage[margin=1in]{geometry} 

\usepackage{amsmath}
\usepackage{amsthm}
\usepackage{amssymb}

\usepackage[utf8]{inputenc}
\usepackage{hyperref}

\usepackage[sort&compress,numbers,square]{natbib}
\usepackage{graphicx, color}
\graphicspath{{figs/}}

\usepackage{algorithm, algpseudocode} 
\usepackage{mathrsfs} 
\usepackage{gensymb}
\usepackage{subfigure}
\usepackage{booktabs}

\title{Unsupervised Severe Weather Detection Via Joint Representation Learning Over Textual and Weather Data}
\author{Athanasios Davvetas$^1$ \and Iraklis A. Klampanos$^1$}

\date{
	$^1$National Centre for Scientific Research ``Demokritos'' \\ Institute of Informatics and Telecommunications \\ Athens \\ Greece \\ \texttt{\{tdavvetas, iaklampanos\}@iit.demokritos.gr}\\
}

\begin{document}
	\maketitle
	
	\begin{abstract}
	    When observing a phenomenon, severe cases or anomalies are often characterised by deviation from the expected data distribution. However, non-deviating data samples may also implicitly lead to severe outcomes. In the case of unsupervised severe weather detection, these data samples can lead to mispredictions, since the predictors of severe weather are often not directly observed as features. We posit that incorporating external or auxiliary information, such as the outcome of an external task or an observation, can improve the decision boundaries of an unsupervised detection algorithm. In this paper, we increase the effectiveness of a clustering method to detect cases of severe weather by learning augmented and linearly separable latent representations.We evaluate our solution against three individual cases of severe weather, namely windstorms, floods and tornado outbreaks. 
	\end{abstract}
    \section{Introduction}
    Anomalies occur in the majority of datasets. They are fairly rare and are often challenging to detect in an unsupervised setting. Due to their lower frequency, the majority of normal samples introduces implicit bias that results in biased predictions. From an unsupervised perspective, one can assume that these rare occurrences can be observed in the outliers of the data distribution. Yet, depending on the application, searching for samples that deviate from the expected data distribution may not improve the detection of an unsupervised method.
    
    In some applications, the occurrence of anomalies might be expected or it may not be trivial to detect deviation from the observed data distribution. An example of such application is detecting cases of severe weather. A heavy rain or windstorm may be considered as normal, depending on the geographic region or the season, etc. These otherwise normal circumstances may lead to natural disasters, even in costly damages or fatalities. However, they can not always be predicted by observing a physical quantity.
    
    To predict these types of occurrences, we need to incorporate external or auxiliary information that can effectively augment the observable features. In this paper, we investigate the effects of incorporating external information in the form of an auxiliary task outcome. We achieve this by utilising a deep learning method called ``Evidence Transfer'' that incrementally manipulates the latent representations of an autoencoder according to external categorical evidence \cite{Davvetas2019}. Evidence transfer allows for joint representation learning, based on external categorical evidence retrieved from textual sources and weather re-analysis data. Evidence transfer successfully manipulates the initial learned representations, resulting in increasead effectiveness during individual severe weather detection.
    
    
    \section{Data and Methods}
    \subsection{Weather Re-analysis Data}
    ERA-Interim \cite{ERA-Interim} re-analysis data are produced with a sequential data assimilation scheme during which prior information from a forecast model is combined with the available observations in order to estimate the state of the global atmosphere, allowing for a better description of past atmospheric conditions. Weather re-analysis data are gridded data depicting atmospheric variables in various timestamps and pressure levels (gravity-adjusted height), leading to 4D variables. They cover a time period of up to 40 years, with less than 1$\degree$ spatial fineness and 6-hour temporal resolution for global region. 
    
    In our experiments, we used ERA-interim data covering the time period from January 1 1979 to May 31 2018 with 6-hour temporal resolution (retrieved from the Research Data Archive of National Center for Atmospheric Research in Boulder, Colorado\footnote{\url{http://rda.ucar.edu/datasets/ds627.0/}}). The spatial resolution is $\approx0.7\degree\times0.7\degree$, containing atmospheric variables across $37$ vertical pressure levels ranging from 1$hPa$ to 1000$hPa$. We reduce the region of gridded data from global region to a Cartesian domain that covers Europe. In order to reduce the domain of our data we used the pre-processor of Weather Research and Forecast (WRF) Model \cite{WRF}, named WPS. The new spatial resolution of our data is of $64\times64$ cells of $75km\times75km$ in the west-east and south-north axes.
    
    In our study, the atmospheric variable of interest is the geopotential height (GHT) which can be seen as a gravity-adjusted height. GHT is often used for its predictive properties \cite{Turkes2002,Krishnamurti2003,Krinitskiy2019}, as well as, to extract weather patterns for other downstream tasks \cite{Klampanos2017}. Severe weather can be predicted via sequences of patterns in the geopotential height (e.g. a cyclone can be observed as a circular pattern). To highlight useful high-level features, such as circular shapes and edges, we extract embeddings through a pre-trained VGG-16 network on ImageNet.  We feed the VGG-16 network with 3 different levels of GHT = 500, 700 and 900 $hPa$ in similar fashion to using the RGB channels of an image. Therefore, a single data sample of shape $3\times64\times64$ is transformed into an embedding of $64\times64$, resulting in a total of 4096 features.
    
    \subsection{Textual Evidence}
    We augment the weather-based embeddings by making use of textual evidence for historic severe weather events, found in Wikipedia. For example, to find severe heavy rain occurrences we search for recorded floods. We extract categorical evidence from textual sources of Wikipedia pages which associate a date to a severe weather event. 
    
    For our experiments we extract the following cases of extreme events in Europe: (1) costly or deadly hailstorms, (2) floods, (3) tornadoes and tornado outbreaks and (4) severe windstorms.
    
    Each of these event types is treated as a binary classification task for predicting a specific severe weather case. The occurence date is used to both reference the weather re-analysis data, as well as the individual tasks. Since the events listed in Wikipedia do not typically supply exact times, we label the whole day of reference as severe, therefore, the minimum span of an event is one day or four 6-hour samples (we remind that the weather re-analysis data are provided in 6-hour increments). 
    
    For each of the aforementioned lists we extract the following fields (for simplicity purposes, ``Event'' is used to represent each individual case of severe weather): Event Name, Event Type, Affected Countries, Location, Country Coordinates (Latitude), Country Coordinates (Longitude), Event Description. The fields regarding the event (name, type, location, description) are extracted from the Wikipedia pages, while the coordinates are retrieved from querying the GeoNames \footnote{\url{https://www.geonames.org}} API. For the majority of extracted events, country names are used to reference the spatial extension of an event which is stored in the ``Affected Countries'' field. More detailed spatial information such as city names or state names are stored in the ``Location'' field when they are available. 
    
    \subsection{Evidence Transfer}
    Evidence transfer \cite{Davvetas2019} is a deep learning method that incrementally manipulates the latent representations of an autoencoder according to external categorical evidence. In the context of evidence transfer, any categorical variable can be utilised as evidence. The most straight forward case of evidence is using the outcome of an auxiliary task. Evidence transfer has been developed with the notion that in practice the availability of external data is either not guaranteed, or we may observe the outcome of external processes without having explicit access to the corresponding dataset. It is a generic method for combining external evidence in the process of representation learning. It makes no assumptions regarding the nature or source of external evidence. It is effective when introduced with meaningful evidence, robust against non-corresponding evidence and modular due to its transfer learning nature.
    
    Evidence transfer is a two step method. During the initialisation step, an autoencoder is trained to reconstruct the input data of the primary task. To ensure robustness, an intermediate step is required. During the intermediate step we train a small biased evidence autoencoder to reconstruct each categorical evidence source. We call the evidence autoencoder, ``biased'' due to introduced limitation in the amount of iterations. Meaningful evidence is able to converge for small amount of iterations, leading to a latent projection of the evidence, however non-corresponding evidence is not able to generalise and therefore produce a uniform-like distribution. During severe weather case detection, we avoid this step, since we know that textual evidence is retrieved from meaningful sources. 
    
    During the transfer step, the initial latent representation are manipulated according to external evidence through the joint optimisation of reconstructing the input, as well as, reducing the cross entropy between an extended softmax layer of the latent space and the external evidence. The loss function of the initialisation step is shown in Equation \ref{eq:init}. In Equation \ref{eq:evitransf} we show the evidence transfer step loss, where $V$ is the set of categorical evidence sources and $Q$ are the extended softmax layers. Structural Similarity Index (SSIM) is used as reconstruction loss function in order to retain the structural information of the data.
    \begin{equation}
      \label{eq:init}
          \ell_{AE} = \mathcal{L}({X}, {X'}) = \frac{1}{N} \sum_{i=1}^{N} SSIM(x^{(i)}, x'^{(i)})
    \end{equation}
    
    \begin{equation}
      \label{eq:evitransf}
          \ell_{EviTransf} = \ell_{AE} + \lambda * \frac{1}{K} \sum_{j=1}^{K}H(V_j, Q_j)
    \end{equation}
    
    \begin{table}
    \caption{Experimental evaluation of evidence transfer for individual severe weather case detection.}
    \label{tab:allone} %
    \subtable{
        \begin{tabular}{ccc}
        \toprule 
        \multicolumn{3}{c}{Windstorm (Baseline)}\tabularnewline
        \midrule 
        Metric & Flood & Tornado\tabularnewline
        \midrule
        Precision & 0.61 & 0.66\tabularnewline
        Recall & 0.71 & 0.87\tabularnewline
        F1-Score & 0.66 & 0.75\tabularnewline
        \midrule 
        \multicolumn{3}{c}{Windstorm (Evidence Transfer)}\tabularnewline
        \midrule
        Metric & Flood & Tornado\tabularnewline
        \midrule
        Precision & 0.84 (+0.23) & 0.79 (+0.13)\tabularnewline
        Recall & 0.74 (+0.03) & 1.00 (+0.13)\tabularnewline
        F1-Score & 0.79 (+0.13) & 0.88 (+0.13)\tabularnewline
        \bottomrule
        \end{tabular}
    }
        \subtable{
        \begin{tabular}{ccc}
        \toprule 
        \multicolumn{3}{c}{Flood (Baseline)}\tabularnewline
        \midrule 
        Metric & Windstorm & Tornado\tabularnewline
        \midrule
        Precision & 0.49 & 0.61\tabularnewline
        Recall & 0.50 & 0.57\tabularnewline
        F1-Score & 0.49 & 0.59\tabularnewline
        \midrule 
        \multicolumn{3}{c}{Flood (Evidence Transfer)}\tabularnewline
        \midrule
        Metric & Windstorm & Tornado\tabularnewline
        \midrule
        Precision & 0.68 (+0.19) & 0.72 (+0.11)\tabularnewline
        Recall & 0.92 (+0.42) & 0.69 (+0.12)\tabularnewline
        F1-Score & 0.78 (+0.29) & 0.71 (+0.12)\tabularnewline
        \bottomrule
        \end{tabular}
    }
        \subtable{
        \begin{tabular}{ccc}
        \toprule 
        \multicolumn{3}{c}{Tornado (Baseline)}\tabularnewline
        \midrule 
        Metric & Windstorm & Flood\tabularnewline
        \midrule
        Precision & 0.26 & 0.24\tabularnewline
        Recall & 0.62 & 1.00\tabularnewline
        F1-Score & 0.36 & 0.38\tabularnewline
        \midrule 
        \multicolumn{3}{c}{Tornado (Evidence Transfer)}\tabularnewline
        \midrule
        Metric & Windstorm & Flood\tabularnewline
        \midrule
        Precision & 0.32 (+0.06) & 0.28 (+0.04)\tabularnewline
        Recall & 0.98 (+0.36) & 0.69 (-0.31)\tabularnewline
        F1-Score & 0.49 (+0.13) & 0.40 (+0.02)\tabularnewline
        \bottomrule
        \end{tabular}
    }
    \end{table}
    
    \begin{table}
    \caption{Experimental evaluation of evidence transfer for severe weather case detection with three sampling strategies.}
    
    \label{tab:TT-OUC-full} %
    \begin{tabular}{cccc}
    \toprule 
    \multicolumn{4}{c}{Baseline}\tabularnewline
    \midrule 
    Metric & Oversample & \textbf{Undersample} & Combine\tabularnewline
    \midrule
    Precision & 0.51 & 0.53 & 0.51\tabularnewline
    Recall & 0.51 & 0.53 & 0.51\tabularnewline
    F1-Score & 0.51 & 0.53 & 0.51\tabularnewline
    \midrule 
    \multicolumn{4}{c}{Evidence Transfer}\tabularnewline
    \midrule
    Metric & Oversample & \textbf{Undersample} & Combine\tabularnewline
    \midrule
    Precision & 0.59 (+0.08) & 0.82 (+0.29) & 0.55 (+0.04)\tabularnewline
    Recall & 0.59 (+0.08) & 0.82 (+0.29) & 0.55 (+0.04)\tabularnewline
    F1-Score & 0.59 (+0.08) & 0.82 (+0.29) & 0.55 (+0.04)\tabularnewline
    \end{tabular}
    \end{table}
    
    \subsection{Class Balancing}
    In our experiments, the original data consist of 57584 weather re-analysis samples in 6-hour increments, while the total amount of severe weather samples without duplicate dates are only 3136 (less than 6\% of the samples). To deal with imbalanced learning, we experiment with three different sampling strategies: (1) Over-sampling the minority class, (2) Under-sampling the majority class, (3) Combination of Over-sampling and Under-sampling
    To over-sample the minority class, we use SMOTE \cite{Chawla2002}. SMOTE generates minority class samples by joining the line segments of k-nearest neighbors. To under-sample the minority class, we perform random under-sample, although more sophisticated under-sampling methods such as the ENN \cite{Wilson1972} (removes data samples that deviate from the majority of k-nearest neighbors) can also be used. A combination of both strategies can be achieved by combining the over-sampling with under-sampling, such as the SMOTEENN method \cite{Batista2004}.
    
    In order to test the effectiveness of each sampling strategy we experiment with using the primary task of learning representations to detect severe weather samples by combining all severe cases into a single class. We manipulate the initial learned space by incorporating the ground-truth labels (i.e the binary task labels of predicting severe from non-severe weather samples). Incorporating evidence that exactly replicates the outcome of the primary task is not realistic, however we use this scenario in order to investigate the best choice of sampling strategy without introducing implicit uncertainty from the choice of external categorical evidence. However, to test its generalisation, we split the ground-truth into train and test and only use the evidence labels during training with evidence transfer.
    
    Quantitative evaluation with the micro average of precision, recall and F1 score metrics for the full dataset (train and test) are presented in Table \ref{tab:TT-OUC-full}. Our experiments indicate that under-sampling the majority class is the most fitting for our case. By reducing the redundancy in the majority class, evidence transfer can more effectively manipulate the initial representations. It additionally allows the linear separation into two classes, while during over-sampling and combine, the implicit bias overcomes the latent space by resulting in a single inseparable cluster.
    
    \subsection{Method overview}
    For all of our experiments, we follow the training procedure of evidence transfer. First, we train a denoising stacked autoencoder to reconstruct the primary task dataset, i.e. the weather re-analysis data. The initialisation step is completely unsupervised, no labels are used during this step. We consider an initial solution to our primary task, the ``baseline'' solution, during which we perform an unsupervised detection method on the initially retrieved latent representations. We perform the same unsupervised detection method on incrementally manipulated latent representations from evidence transfer in order to compare its effectiveness. We supply the additional evidence sources based on the textual severe weather dataset.
    
    During experimental investigation of the best sampling strategy, One Class SVM was used as an unsupervised detection method. For the cases of detecting individual severe weather cases we use $k$-means clustering with $k$=2 as an unsupervised detection method, except a single case where agglomerative clustering was used instead (ground-truth: windstorm, evidence: tornado).
    
    \section{Experimental Evaluation}
    We experiment with individually detecting windstorms, floods and tornado outbreaks. We avoid using the hail events due to limited amount of samples. We rotate between the different severe cases by selecting one case as the ground truth and alternate between using the rest as external evidence. For example, we select windstorm weather samples and a portion of non-severe samples as our primary task -- ground truth, while another case, e.g flood, is selected as the auxiliary task -- external evidence. We further under-sample the remaining non-severe weather cases in order to match the number of severe weather samples. 
    
    In Table \ref{tab:allone}, we report experimental results in terms of precision, recall and F1-score for the anomalous class. Introducing external evidence leads to linearly separable representations that increase the effectiveness of clustering, and therefore detecting the severe weather samples. Even though evidence transfer is a scalable method that can use multiple sources of evidence, in this case, it is not as effective, due to ground truth and external evidence contradicting each other for some portion of the data samples.
    
    In our experiments, the final dataset consists of non-severe samples ($\approx$500 after under-sampling to balance the individual severe class), one severe class as the primary task or ground truth and one as the external evidence. As an example, consider the task of predicting windstorm samples as ground truth and the task of predicting flood samples as external evidence. For the task of predicting windstorms, non-severe samples and flood samples are labelled as ``normal''. However, for the task of predicting floods, non-severe samples and windstorm samples are labelled as ``normal''. Therefore, external evidence contradicts the ground truth during non-severe samples. Introducing more sources of external evidence increases the contradiction for non-severe samples, leading to increased uncertainty during clustering.
    
\begin{figure}
      \centering
    \subfigure[Baseline of "Windstorm - Flood" Combination]{\includegraphics[width=.45\textwidth]{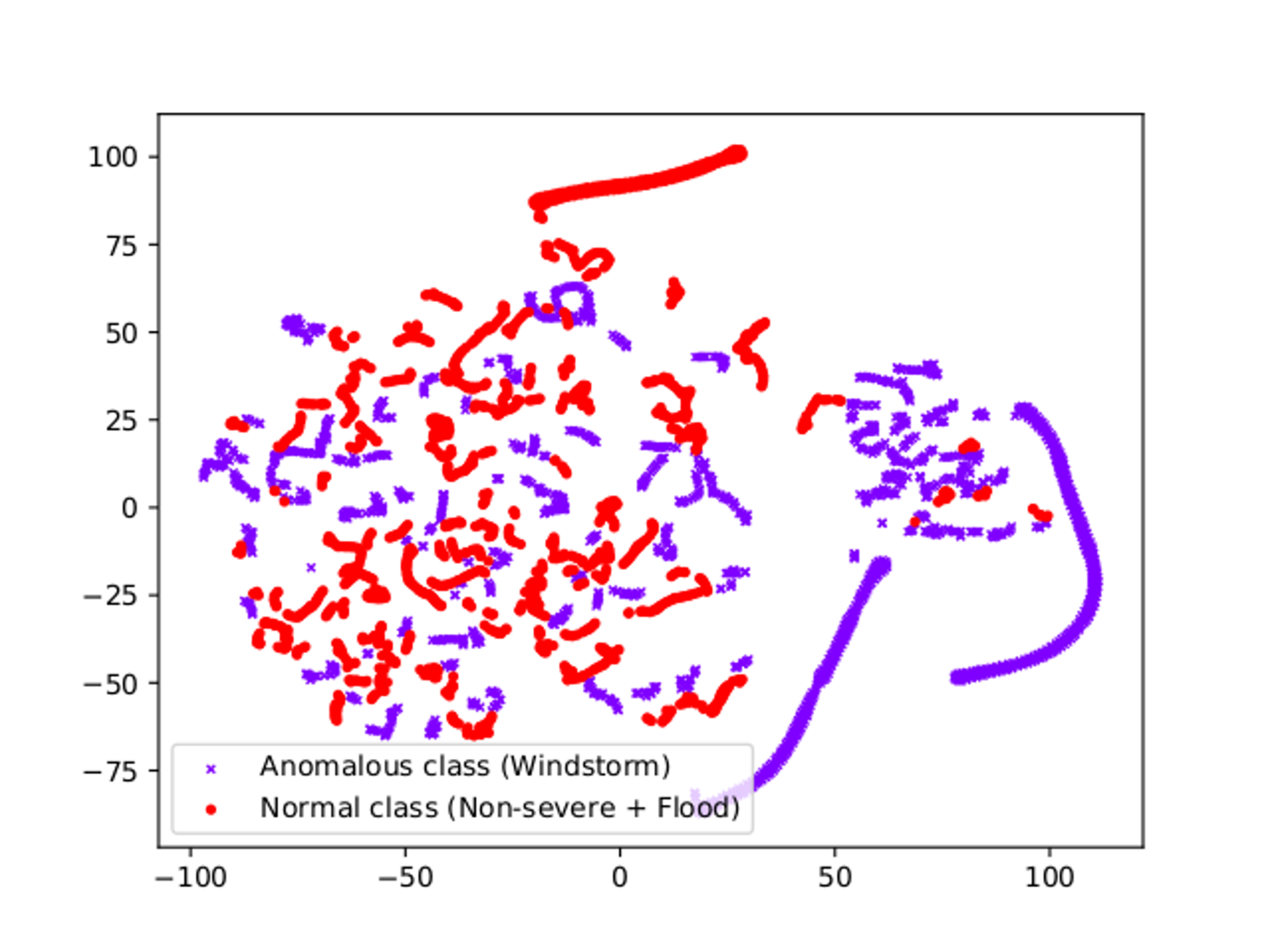}}
    \subfigure[Evidence transfer combination of "Windstorm - Flood"]{\includegraphics[width=.45\textwidth]{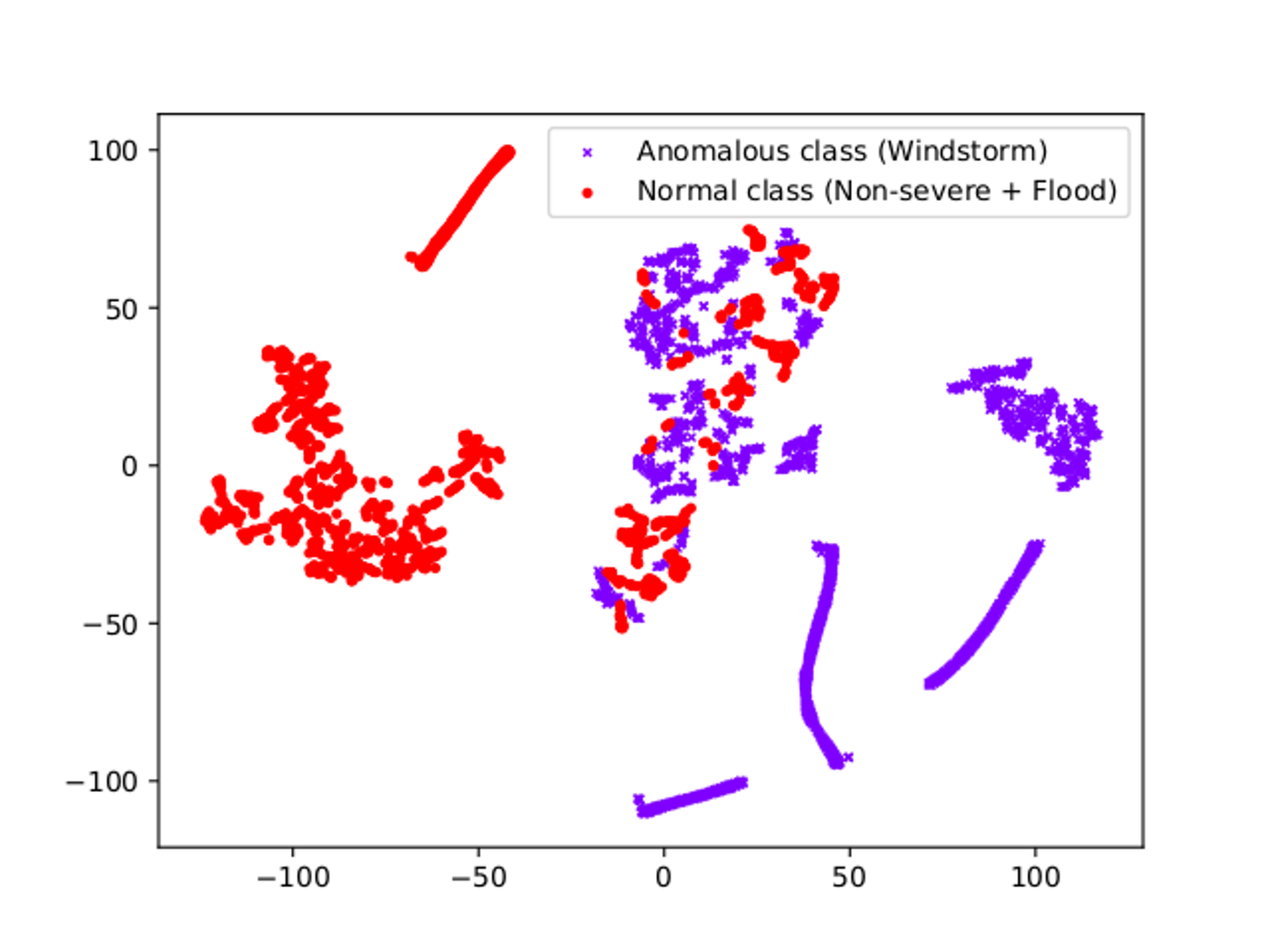}}
      \caption{t-SNE 2d projections of the initial and Evidence Transfer representations of originally 10 features. The initial latent space consists of a "mixed" cluster that can be seen as a single class in an unsupervised setting. However, after evidence transfer, the latent representations are linearly separable allowing for improved decision boundaries.}
      \label{fig:windf}
    \end{figure}

    However, both quantitatively, as shown in Table \ref{tab:allone}, as well as qualitatively (ground-truth: windstorm, evidence: flood, depicted in Figure \ref{fig:windf}) introducing a single source of evidence improves the outcome of clustering method by pushing the latent representations to becoming linearly separable and therefore improving the effectiveness for both $k$-means and agglomerative clustering. 
    
    \section{Future Work and Conclusions}
    In this paper, we investigated using evidence transfer to improve a primary task of detecting individual cases of severe weather. By incorporating auxiliary tasks extracted from textual sources, we effectively manipulated the latent space of an autoencoder using evidence transfer, in order to increase the effectiveness of severe weather detection. Making latent representations incrementally linearly separable resulted in improving the effectiveness of $k$-means and agglomerative clustering. Additionally, we investigated the best sampling method for our imbalanced class of detecting severe cases with non-observable predictors, by evaluating the effectiveness of evidence transfer in one-class SVM (with linear kernel) prediction.
    
    Future work is directed towards utilising the temporal aspect of weather re-analysis data. For our experiments, we mostly focused on using embeddings extracted from an image recognition task. However, retrieving temporally-aware embeddings from raw data, e.g. via a recurrent autoencoder, could improve the individual detection of severe weather cases by exploiting the temporal aspect of the data.
    Additionally, since the under-sampling strategy appears to perform better for this problem, it would be beneficial to increase the total amount of severe weather samples from additional sources.

	\paragraph{Acknowledgements} 
	This work has been supported by the Industrial Scholarships program of Stavros Niarchos Foundation.

	\bibliography{main}

\end{document}